\documentclass[letterpaper, 10 pt, conference]{ieeeconf} 

\IEEEoverridecommandlockouts
\usepackage{balance}
\usepackage{graphicx}
\usepackage{amsfonts} 
\usepackage{amsmath}
\usepackage{booktabs}
\usepackage{multirow}
\usepackage{hyperref}
\usepackage{amssymb}
\usepackage{mathtools}
\usepackage{url}
\usepackage{subcaption}
\usepackage{color}
\usepackage{xcolor}
\usepackage{marvosym}

\title{MLA: A Multisensory Language-Action Model for Multimodal Understanding and Forecasting in Robotic Manipulation}

\author{
Zhuoyang Liu$^{1*}$,
Jiaming Liu$^{1*\dagger}$,
Jiadong Xu$^{1}$,
Nuowei Han$^{1}$,
Chenyang Gu$^{1}$,
Hao Chen$^{3}$,
Kaichen Zhou$^{1}$, \\
Renrui Zhang$^{3}$,
Kai Chin Hsieh$^{1}$,
Kun Wu$^{2}$,
Zhengping Che$^{2\dagger}$,
Jian Tang$^{2}$,
Shanghang Zhang$^{1}\textsuperscript{\Letter}$%
\thanks{$^*$Equal contribution. $^\dagger$Project lead. $^{1}$State Key Laboratory of Multimedia Information Processing, 
School of Computer Science, Peking University. $^{2}$Beijing Innovation Center of Humanoid Robotics (X-Humanoid). $^{3}$The Chinese University of Hong Kong (CUHK).}
}

\begin{document}

\maketitle
\thispagestyle{empty}
\pagestyle{empty}

\begin{abstract}
Vision-language-action models (VLAs) have shown generalization capabilities in robotic manipulation tasks by inheriting from vision-language models (VLMs) and learning action generation.
Most VLA models focus on interpreting vision and language to generate actions, whereas robots must perceive and interact within the spatial-physical world.
This gap highlights the need for a comprehensive understanding of robotic-specific multisensory information, which is crucial for achieving complex and contact-rich control.
To this end, we introduce a multisensory language–action (MLA) model that collaboratively perceives heterogeneous sensory modalities and predicts future multisensory objectives to facilitate physical world modeling.
Specifically, to enhance perceptual representations, we propose an encoder-free multimodal alignment scheme that innovatively repurposes the large language model itself as a perception module, directly interpreting multimodal cues by aligning 2D images, 3D point clouds, and tactile tokens through positional correspondence.
To further enhance MLA’s understanding of physical dynamics, we design a future multisensory generation post-training strategy that enables MLA to reason about semantic, geometric, and interaction information, providing more robust conditions for action generation.
For evaluation, the MLA model outperforms the previous state-of-the-art 2D and 3D VLA methods by 12\% and 24\% in complex, contact-rich real-world tasks, respectively, while also demonstrating improved generalization to unseen configurations.
Project website: \url{https://robotic-mla.github.io/}

\end{abstract}

\section{INTRODUCTION} 

Recent robot imitation learning has achieved remarkable advances in training policies from expert demonstrations to perform diverse vision–language manipulation tasks. Meanwhile, vision–language models (VLMs)~\cite{karamcheti2024prismatic, wang2024qwen2} pre-trained on internet-scale data have been proven to possess strong capabilities in common-sense reasoning in general scenarios. Building on these progresses, vision–language–action (VLA) models have been proposed~\cite{black2024pi_0,liu2025hybridvla,kim2024openvla}, which not only inherit the properties of VLMs but also extend them by training with robot demonstrations for action prediction. As a result, VLA models demonstrate impressive generalization and precise manipulation, effectively mapping human instructions and visual observations to the robot control signal.

\begin{figure}[t]
\includegraphics[width=0.48\textwidth]{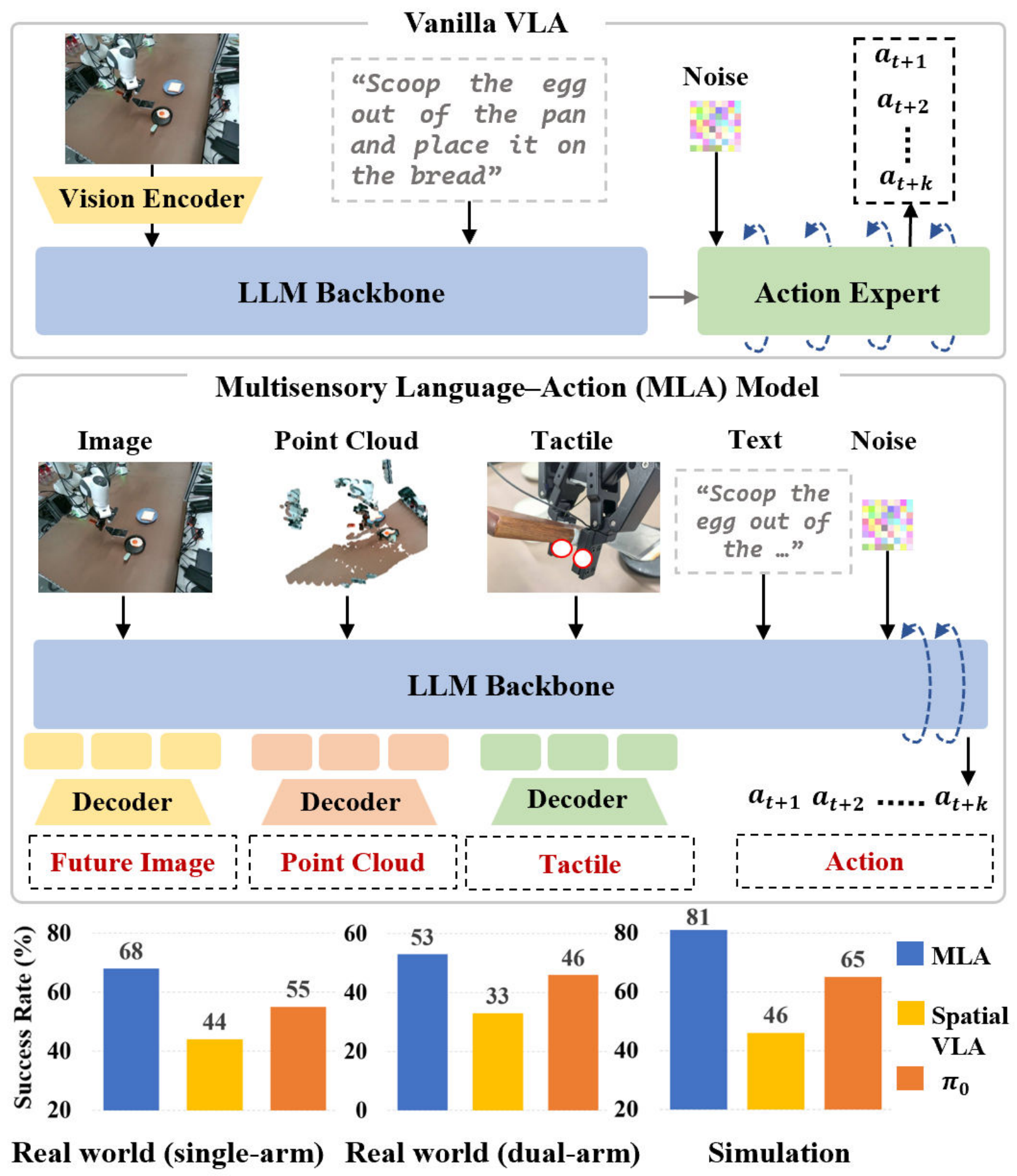}
\centering
\vspace{-0.45cm}
\caption{(a) Unlike vanilla VLA methods that rely on 2D images and natural language instructions to generate actions, (b) we propose MLA, a multisensory language–action model that collaboratively processes diverse robotic-specific modalities and predicts their future states to enhance physical dynamics modeling in robotic control. (c) MLA achieves state-of-the-art performance across a variety of real-world and simulation tasks.
}
\label{fig:intro}
\vspace{-0.65cm}
\end{figure}

In the real world, robots must perceive spatial environments, reason about semantic relationships, and interact with dynamic environment configurations. However, most existing VLA models rely primarily on 2D image integration~\cite{li2024cogact,black2024pi_0}, which is fundamentally inadequate for capturing spatial dependencies and modeling physical dynamics.
\textbf{On the one hand}, to address these limitations, several studies enhance VLAs with richer multimodal observations. Specifically, some approaches incorporate 3D inputs to improve geometric scene understanding~\cite{zhen20243d,qu2025spatialvla}, while others introduce tactile signals to capture interaction feedback from manipulated objects~\cite{cheng2025omnivtla,huang2025tactile}.
Existing VLA models often require modality-specific encoders to enrich perceptual capacity, which undermines efficiency. Furthermore, without pre-training on multisensory inputs, the large language model (LLM) backbone of VLAs exhibits limited representation to align with the newly introduced multimodal features.
\textbf{On the other hand}, several VLA studies attempt to reason about the physical dynamics by predicting future states, such as subgoal images and camera-view depth maps~\cite{zhang2025dreamvla,wang2025unified,zhao2025cot}.
However, these approaches remain limited in predicting complete point cloud structures and tactile interaction information, which are essential not only for understanding complex, contact-rich scenes but also for effective motion planning in robotic manipulation.
Consequently, a critical question arises: how can multisensory modalities be integrated into a unified representation and predicted in their future states to collaboratively enhance VLA models’ physical-world understanding and action generation?

To address this question, we propose \textbf{MLA}, a multisensory language–action model that collaboratively processes diverse sensory inputs and predicts their corresponding future states to enhance physical-world modeling for robotic control.
As shown in Figure~\ref{fig:intro}, to avoid introducing additional modality-specific encoders that lack pretraining alignment with LLM’s embeddings, MLA adopts an encoder-free multimodal alignment mechanism, repurposing the initial transformer blocks of the LLM as a perception module to directly interpret visual, geometric, and tactile cues.
In particular, we project 3D points and the spatial positions of the tactile gripper onto 2D image planes using camera parameters, thereby constructing cross-modal positional mappings. These positional correspondences serve as positive pairs for token-level contrastive learning, aligning multimodal features within the LLM’s embedding space.
This position-guided consistency constraint enhances the multimodal representations of our MLA model and supports more comprehensive physical-world perception.
To further enhance the LLM’s understanding of physical robotic scenes, we propose a future multisensory generation post-training strategy. 
Specifically, the lightweight transformer-based decoders and tailored generation scheme are designed to process the LLM’s final-layer features and generate the future states of multiple modalities, including 2D images, 3D point clouds, and tactile signals.
Through this predictive process, MLA is able to reason about physical dynamics from multiple dimensions, encompassing semantic information, geometric structures, and object-centric interactions.
Notably, the proposed methods are applied only during training and do not affect MLA’s inference efficiency, while enriching feature conditions for action generation.

Since existing open-source real-world datasets~\cite{open_x_embodiment_rt_x_2023,khazatsky2024droid,wu2024robomind} lack multisensory information, we pretrain the LLM solely on large-scale image–action paired datasets following common practice~\cite{kim2024openvla,liu2025hybridvla}, including more than 570K trajectories. 
Subsequently, we perform supervised fine-tuning (SFT) on downstream task datasets using the proposed encoder-free multimodal alignment mechanism, and finally conduct future multisensory generation post-training, progressively equipping our model with the ability to integrate perception, understanding, and action generation from multisensory inputs in the real physical world.
To systematically evaluate our model, we design six complex, contact-rich real-world robotic experiments covering both single- and dual-arm manipulation tasks, where MLA achieves state-of-the-art success rates and demonstrates strong generalization to unseen objects and backgrounds. For reproducibility, we further evaluate MLA on the RLBench~\cite{james2020rlbench} simulator and also obtain competitive performance. As tactile sensing in simulation is not realistic, we incorporate tactile signals only in real-world experiments. Our contributions are summarized as follows:

\begin{itemize}

\item We propose MLA, a multisensory language-action model with an encoder-free multimodal alignment mechanism, repurposing the LLM itself to directly align with and interpret image, point cloud, and tactile cues.

\item To further strengthen MLA’s understanding of physical dynamics, we introduce a future multisensory generation post-training strategy that enables it to reason about semantic, geometric, and interaction information, providing more robust conditions for action generation.

\item Through a progressive pipeline of pretraining, SFT, and post-training, MLA achieves state-of-the-art success rates and strong generalization on complex real-world tasks, including both single- and dual-arm manipulation.

\end{itemize}

\section{RELATED WORK}

\textbf{Vision-language-action (VLA) models} \cite{kim2024openvla,black2024pi_0,li2024cogact,qu2025spatialvla,liu2025hybridvla} have advanced rapidly with the development of vision-language models (VLMs) and large-scale robotic datasets.
PaLM-E~\cite{driess2023palm} pioneered the adaptation of VLMs to robotic data, and subsequent works followed this paradigm, further extending its capabilities \cite{kim2024openvla, black2024pi_0}.
Meanwhile, diffusion and flow modeling have emerged as effective tools for modeling the multimodal distributions of robotic actions \cite{chi2023diffusionpolicy,ze20243d}. This has motivated approaches that condition continuous action experts on VLM representations \cite{li2024cogact,intelligence2025pi05visionlanguageactionmodelopenworld,black2024pi_0}, as well as recent scaling efforts using transformer-based diffusion architectures~\cite{liurdt,chen2025fast,liu2025hybridvla}. Moreover, some studies further enhance action generation by incorporating richer sensory inputs, such as 3D point clouds \cite{qu2025spatialvla,li2025pointvlainjecting3dworld} and tactile signals \cite{cheng2025omnivtla,huang2025tactile}.
These methods often require modality-specific encoders which undermines efficiency. Also, without multisensory pre-training, the LLM backbone struggles to align and interpret them efficiently.

\textbf{Robotic world knowledge forecasting policy}, which predicts and reasons about future world knowledge, has gained considerable attention in robotics for its ability to capture the dynamics of the physical environment.
Early attempts~\cite{hu2024video,zhou2024robodreamer} employed generative models to directly predict future images, and then leveraged the learned representations to train an action generator.
Subsequent work~\cite{zhang2025dreamvla,bu2025agibot,liu2026last0latentspatiotemporalchainofthought} explored the use of latent action tokens as forward-dynamics representations for action planning and generation. Another line of VLA research~\cite{wu2023unleashing,zhao2025cot,zhang2025upvlaunifiedunderstandingprediction,wang2025unified} focuses on leveraging future state prediction to facilitate action generation.
While these methods are confined to 2D image prediction and struggle with complex, contact-rich scenes, MLA introduces comprehensive multisensory forecasting for robotics, yielding more robust representations for action generation.

\begin{figure*}[t]
    \centering
    \includegraphics[width=\textwidth]{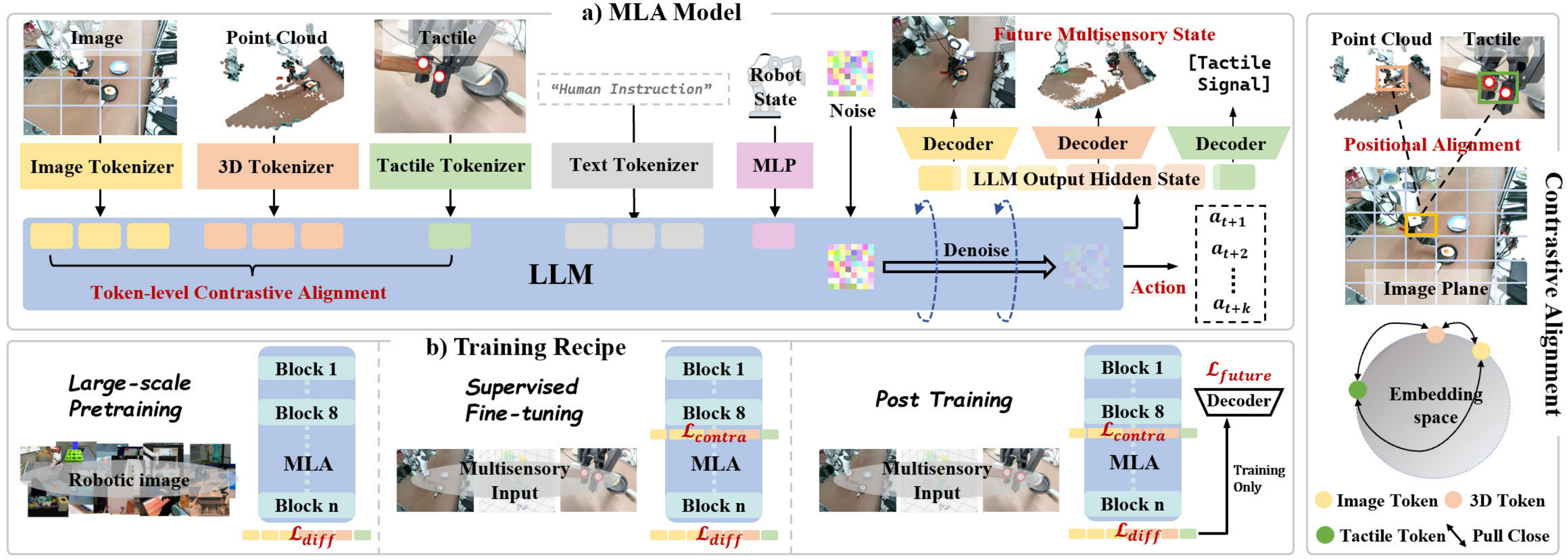}
    \vspace{-0.4cm}
    \caption{\textbf{Overall Framework of MLA.} a) Beyond language instructions and robot states, MLA introduces an innovative encoder-free multimodal alignment mechanism that directly enables the LLM to integrate RGB images, point clouds, and tactile signals, aligning them through token-level contrastive learning. Furthermore, MLA incorporates a future multisensory generation post-training strategy, allowing the model to generate future multisensory states and providing more robust conditions for action generation. b) MLA adopts a three-stage training paradigm: large-scale pretraining, supervised fine-tuning with cross-modal alignment, and post-training with future state prediction.}
    \label{fig:model}
    \vspace{-0.4cm}
\end{figure*}

\section{METHOD}

\subsection{Preliminary} 
Similar to the VLA problem~\cite{black2024pi_0, li2024cogact}, our MLA imitation learning is formulated as a probabilistic sequence decision-making task. At each timestep $t$, the policy $\pi_\theta$ takes multimodal inputs, including the image observation $I_t$, point cloud $P_t$, tactile signal $T_t$, robot state $S_t$, and language instruction $L$. It then predicts both the immediate action sequence $a_{t:t+H}$ and the future keyframe observations across modalities $I_{t+N}, P_{t+N}, T_{t+N}$. Formally, the generative process is expressed as:
$$a_{t:t+H}, I_{t+N}, P_{t+N}, T_{t+N} \sim \pi_\theta \big( \,\cdot \mid I_t, P_t, T_t, S_t, L \big).$$
 
We follow the experimental setup with Franka single- and dual-arm configurations and represent actions as end-effector pose~\cite{kim2024openvla,liu2025hybridvla}. In the single-arm setting, each action is a 7-DoF vector $a_t = (\Delta x, \Delta y, \Delta z, R_r, R_p, R_y, g)$, where $\Delta x, \Delta y, \Delta z$ denote the Cartesian position delta, $R_r, R_p, R_y$ correspond  to the Euler angles for rotation, and $g$ is the gripper width. In the dual-arm configuration, the action is represented by concatenating two 7-DoF vectors into a 14-DoF representation.

\subsection{MLA Architecture}
As shown in Figure~\ref{fig:model}, our proposed MLA model is built upon a LLM, where the parameters are initialized from the LLM backbone of Prismatic VLM~\cite{karamcheti2024prismatic}, similar to prior work~\cite{kim2024openvla}. Distinct from conventional VLA frameworks that employ vision encoders, our model introduces lightweight tokenizers that directly convert raw multisensory inputs into a shared token sequence and repurpose the LLM itself as a unified MLA model. Furthermore, we incorporate transformer-based decoders that predict future multimodal states.

\textbf{Image Tokenizer.} For each input image $I \in \mathbb{R}^{H \times W \times 3}$, our Vision Tokenizer converts it into a compact token sequence. Following previous works~\cite{chen2025sharegpt, wang2024emu3}, the image is divided into non-overlapping patches of size $14 \times 14$, yielding a token sequence of length $N_{\mathrm{img}} = 256$ with a batchsize of $B$ and an embedding dimension of $d_h = 4096$, i.e., $f^{\mathrm{img}} \in \mathbb{R}^{B \times N_{\mathrm{img}} \times d_h}$.

\textbf{3D Point Cloud Tokenizer.} Given raw point clouds $P \in \mathbb{R}^{B\times 1024 \times 3}$, our 3D Tokenizer partitions the points into local groups centered at sampled anchor points. Following \cite{jia2025lift3d}, the tokenizer consists of three blocks, each incorporating farthest point sampling (FPS)~\cite{qi2017pointnet} for downsampling, k-nearest neighbors (KNN) for local aggregation, and a learnable linear layer for feature encoding. After 3D tokenization, we obtain a compact representation consisting of $N_{\mathrm{pc}}=256$ tokens, $f^{\mathrm{pc}} \in \mathbb{R}^{B\times N_{\mathrm{pc}} \times d_h}$.

\textbf{Tactile Tokenizer.} For tactile sensing, we design a simple MLP-based tokenizer to embed low-dimensional tactile signals into the shared token space. Specifically, we attach two tactile sensors to the gripper fingers. From each sensor, we extract six values: normal force, tangential force, and tangential force direction (two components each). 
The raw signal is processed by a lightweight MLP, producing a tactile token $f^{\mathrm{tac}} \in \mathbb{R}^{B\times 1 \times d_h}$.

\textbf{LLM Backbone.} We adopt LLaMA-2 7B as our base model and repurpose it into a unified perception-and-reasoning policy. Specifically, tokens from images, point clouds, tactile signals, and language are projected into a shared embedding space $f \in \mathbb{R}^{B \times N_t \times 4096}$ and jointly processed by the LLM. The noise tokens required by the diffusion-based action head are appended to the end of the token sequence, enabling the model to perform diffusion modeling.
Diffusion noise and timesteps are embedded through MLP projectors.
This design eliminates the need for separate modality-specific encoders and fully leverages the large-scale pretrained LLM to directly interpret robotic-specific multisensory cues and generate robust actions. 

\textbf{Future Prediction Decoder.} For future multisensory generation, we adopt transformer decoders to predict future sensory observations from the LLM’s final hidden states $h$. Each decoder maps the unified multimodal embeddings into its target space, such as image, point cloud, and tactile vectors, and is supervised by the corresponding future state. The transformer-based decoder follows a standard query–key–value attention design, consisting of four stacked self-attention and feed-forward layers, enabling effective modeling of the multimodal embeddings.

\subsection{Encoder-Free Multimodal Alignment}
\label{sec:EMA}

Previous VLA models~\cite{kim2024openvla, black2024pi_0} rely on vision encoders that are large-scale pretrained on general-domain data, such as SigLIP~\cite{zhai2023siglip}, to process robotic observations. However, these encoders are rarely trained on robot-domain datasets and have not been exposed to robotic-specific sensors. As a result, their representations are limited in aligning with and interpreting robotic data. In addition, pretraining newly introduced encoders often incurs substantial computational cost, and their incorporation constrains inference efficiency.
Inspired by prior works on contrastive learning~\cite{zhai2023siglip}, which adopt a self-supervised approach to align semantic information from heterogeneous modalities into a unified embedding space, we propose an \textit{Encoder-Free Multimodal Alignment} method. 
This method repurposes the initial transformer blocks of the LLM as a unified perception module via token-level contrastive loss, enhancing multisensory representations without the need for additional modality-specific encoders.
In practice, we employ the embedding features from the 8th transformer block for self-supervised alignment and further examine the effect of using different block outputs in our ablation study.

\textbf{Formulation of Positive and Negative Pairs.} 
For the Transformer-based model, the positional indicators of tokens can provide both positional alignment and semantic contextual alignment~\cite{tang2025any2point}. Therefore, we construct cross-multisensory positional mappings to formulate the positive and negative pairs in our token-level contrastive loss. In contrast, directly treating multimodal tokens with misaligned positional information as positive pairs may lead to semantic misalignment.
As shown in the right part of Figure~\ref{fig:model}, we project 3D points and the 3D positions of tactile grippers onto 2D image planes using the camera parameters. Since each 3D point cloud token ($\{f_{i}^{pc}\}_{i=1}^{N_{pc}}$) is aggregated from a set of 3D points, we unproject its center point into 2D image coordinates. For the tactile token (${f^{tac}}$), we directly read the robot state and project the tactile gripper’s position in the 3D world coordinate onto the 2D image plane.
Subsequently, we identify the corresponding 2D image patch onto which these features project, and align the 3D token and tactile token with the corresponding 2D token ($\{f_{j}^{img}\}_{j=1}^{N_{img}}$) as positive pairs ($f_{j}^{img}$–$f_{i}^{pc}$–$f^{tac}$), while the remaining unmatched tokens are treated as negative pairs.

\textbf{Image–Point Cloud Alignment.} Since image and 3D embeddings have the same token sequence length (256), we apply a token-level InfoNCE loss to pull positive pairs together in the embedding space and push negative pairs apart, where $\tau$ denotes the temperature.
{
$$\mathcal{L}_{\mathrm{img\_pc}} = - \frac{1}{256}  \sum_{i=1}^{256} \log \frac{\exp\left(\langle f^{\mathrm{img}}_{j}, f^{\mathrm{pc}}_{i}\rangle / \tau \right)} {\sum_{j=1}^{256} \exp\left(\langle f^{\mathrm{img}}_{j}, f^{\mathrm{pc}}_{i}\rangle / \tau \right)}$$
}

\textbf{Tactile–Image and Point Cloud Alignment.} 
In the single-arm setting, since tactile embeddings consist of a single token, this yields one positive sample per tactile token $(f^{\mathrm{tac}}, f^{\mathrm{img}}_j)$ and $(f^{\mathrm{tac}}, f^{\mathrm{pc}}_i)$, while other tokens serve as negatives. A unidirectional contrastive loss is applied to pull the tactile embedding toward its corresponding tokens:
$$\mathcal{L}_{\mathrm{tac\_img/pc}} = - \log \frac{\exp(\langle f^{\mathrm{tac}}, f^{\mathrm{img/pc}}_{j/i}\rangle / \tau)}{\sum_{j/i=1}^{256} \exp(\langle f^{\mathrm{tac}}, f^{\mathrm{img/pc}}_{j/i}\rangle / \tau)}$$ 
The overall contrastive objective is the sum of the three losses:
$\mathcal{L}_{\mathrm{contrastive}} = \mathcal{L}_{\mathrm{img\_pc}} + \mathcal{L}_{\mathrm{tac\_img}} + \mathcal{L}_{\mathrm{tac\_pc}}$.
Through this contrastive learning objective, the model effectively captures consistent semantic and spatial information, enabling multimodal features to be seamlessly integrated within the LLM’s unified embedding space.

\begin{figure*}[t]
    \centering
    \includegraphics[width=0.97\textwidth]{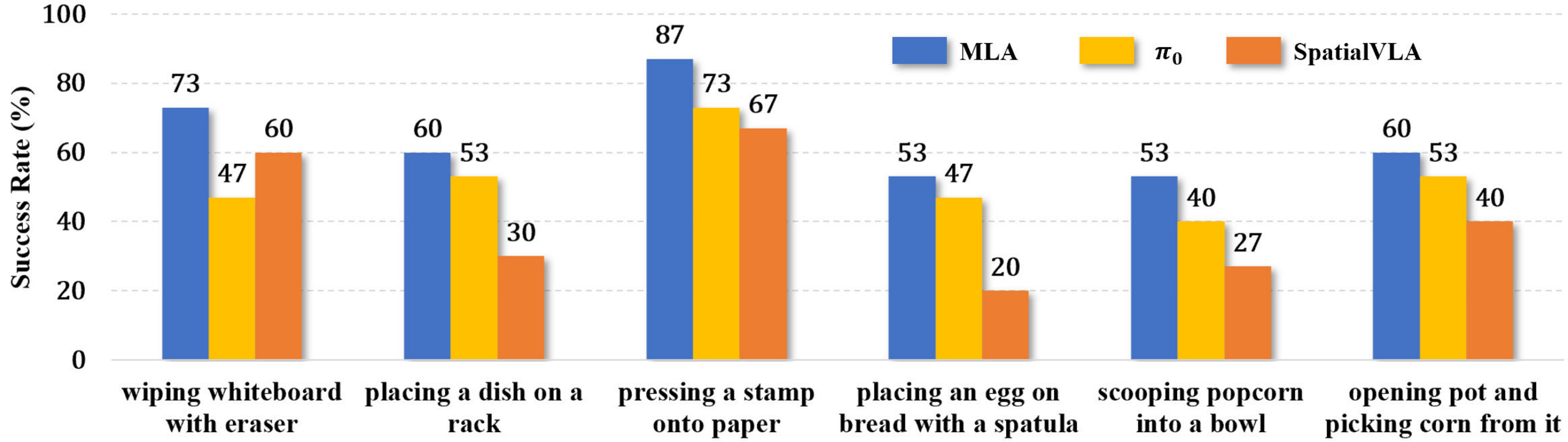}
    \vspace{-0.2cm}
    \caption{\textbf{Real-world results.} All models are evaluated over 15 rollouts from different manipulated object positions on the tabletop, with task completion determined by human judgment.}
    \label{fig:realmain}
    \vspace{-0.4cm}
\end{figure*}

\subsection{Future Multisensory Generation}
\label{sec:FMG}

While some VLA studies~\cite{zhang2025dreamvla, zhang2025upvlaunifiedunderstandingprediction, zhao2025cot} adopt future observation prediction to enable models to reason about physical dynamics, they still fall short in forecasting diverse robotic-specific modalities that are essential for fully capturing the semantic, geometric, and interaction information of the physical world. To further enhance MLA’s understanding of robotic physical scenes, we propose a future multisensory generation post-training strategy, making the first attempt to jointly forecast the future states of images, point cloud, and tactile modalities that are most relevant for manipulation.

\textbf{Image Prediction.} For the visual stream, we adopt a transformer-based decoder, where the last-layer hidden states of the LLM are injected as input features and the future image generation is supervised with an MSE loss. Unlike previous VLA methods that predict dense future frames (close to the current timestep), MLA predicts future keyframes. Following prior work~\cite{jia2025lift3d}, keyframes are identified based on changes in the robotic joint velocity and action transitions. To ease optimization of the image generation loss, background pixels are removed using the corresponding depth map, restricting prediction to foreground regions.

\textbf{Point Cloud Prediction.} For 3D geometry, we adopt a transformer-based decoder to reconstruct the next-keyframe point cloud. Inspired by the masked autoencoder method for point clouds~\cite{pang2022maskedautoencoderspointcloud}, we partition the ground-truth point cloud into $G$ local patches by sampling $G$ center points with FPS and grouping $M$ neighboring points for each center using KNN. The decoder then outputs the predicted coordinates $\hat{P} \in \mathbb{R}^{G \times M \times 3}$, supervised with Chamfer Distance against the ground truth $P$. This operation enhances the stability of future point cloud prediction by first aligning coarse center points to establish the basic 3D structure, and subsequently refining local neighbor points.

\textbf{Tactile Prediction.} For tactile feedback, the decoder outputs a low-dimensional tactile embedding supervised by an MSE loss against the ground truth.

By jointly predicting the future states of images, point clouds, and tactile signals, MLA achieves more comprehensive feature representations across semantic, geometric, and interaction dimensions. It is worth noting that these future-state prediction losses are applied only during the post-training stage and do not affect inference efficiency.

\subsection{Overall Training Recipe}
\textbf{Large-Scale Pretraining.} Similar to previous VLA methods~\cite{kim2024openvla}, we construct a large-scale dataset of over 570K trajectories by combining diverse open-source datasets, such as Open-X-Embodiment~\cite{open_x_embodiment_rt_x_2023} and RoboMIND~\cite{wu2024robomind}. Since the observations in these datasets primarily consist of image and language modalities, we pretrain MLA using only these inputs for 10 epochs. For the other modalities, we reserve their token positions in the sequence, ensuring a smooth transition to subsequent training stages. For the action generation loss ($\mathcal{L}_{\mathrm{diff}}$), we adopt a standard DDPM objective, minimizing the MSE between the predicted and ground-truth noise.

\textbf{Supervised Fine-Tuning.} The pretrained model is subsequently adapted to high-quality, task-specific datasets using the proposed encoder-free multimodal alignment mechanism. In this stage, all multisensory modalities are introduced, including image, point cloud, tactile signals, and language instruction. We incorporate the proposed contrastive loss (as detailed in Section~\ref{sec:EMA}) to enhance MLA’s cross-modal alignment and multimodal representations. The overall training objective loss is
$\mathcal{L}_{\mathrm{sft}} = \mathcal{L}_{\mathrm{diff}} + \mathcal{L}_{\mathrm{contrastive}}$.

\textbf{Post-Training.} Finally, the model undergoes future multisensory generation post-training. In this stage, the training data and input modalities are the same as in the SFT phase. Additionally, the training incorporates future multimodal prediction supervision, as described in Section~\ref{sec:FMG}, enabling the model to capture physical dynamics and thereby achieve more robust action generation. The overall supervision is $\mathcal{L}_{\mathrm{post}} = \mathcal{L}_{\mathrm{diff}} + \mathcal{L}_{\mathrm{contrastive}} + \mathcal{L}_{\mathrm{future}}$. 
Note that we perform SFT followed by post-training to progressively equip our model with the ability to integrate perception, understanding, and action generation from multisensory inputs in the real physical world. During inference, we employ DDIM~\cite{song2020denoising} with $n$ sampling steps (e.g., $n=4$).

\begin{figure}[h]
\includegraphics[width=0.48\textwidth]{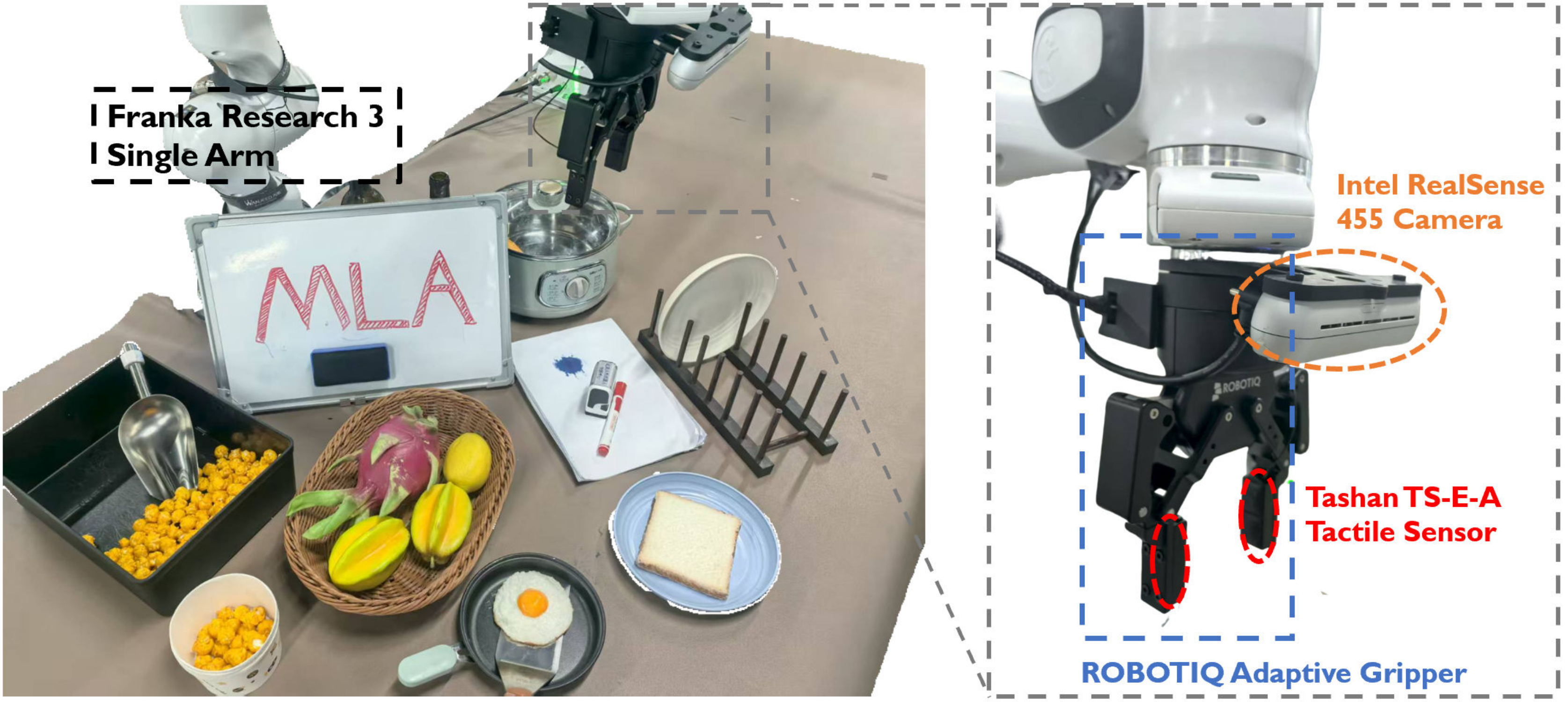}
\centering
\vspace{-0.45cm}
\caption{\textbf{Single-arm Experiment Setup.} We show the details about single-arm setup and assets of real-world experiments.}
\label{fig:singleset}
\vspace{-0.25cm}
\end{figure}

\section{EXPERIMENTS}

In Section~\ref{sec:RWE}, we compare MLA model with recent VLA models on single- and dual-arm real-world tasks. Section~\ref{sec:AS} presents an ablation study of each component, while Section~\ref{sec:GE} demonstrates MLA’s generalization in real-world settings. Section~\ref{sec:SE} benchmarks MLA against VLA baselines in the RLBench simulator for reproducibility.

\subsection{Real-World Experiment}
\label{sec:RWE}
\textbf{Real-World Experiment Setup.} We evaluated four complex contact-rich tasks on a single-arm Franka robot and two tasks on a dual-arm setup combining two Franka robots. As shown in Figure~\ref{fig:singleset}, for the single arm, two RealSense D455 cameras were used to provide image and point cloud data from a third-person view and a wrist view, with only the third-person view contributing to cross-modal alignment. Each gripper was equipped with two tactile sensors (Tashan TS-E-A). For the dual arm, three D455 cameras were employed, including one third-person view and two wrist views.

\textbf{Self-collected Data.} For the single-arm setting, we designed four contact-rich tasks: (1) pressing a stamp onto paper, (2) wiping a whiteboard with an eraser, (3) placing a dish on a rack, and (4) placing an egg on bread with a spatula. For the dual-arm setting, we evaluated two collaborative tasks: (1) scooping popcorn into a bowl and (2) opening a pot lid and picking corn from the pot. All demonstrations were collected using the Gello~\cite{wu2024gellogenerallowcostintuitive} platform, with 200 high-quality demonstrations per task.

\textbf{Training and Evaluation Details.} 
We train MLA for 300 epochs during SFT and 100 epochs during post-training using the AdamW optimizer. Baselines are initialized with their pretrained parameters and fine-tuned under their respective protocols. We compare against two closely related baselines: $\pi_0$~\cite{black2024pi_0}, a state-of-the-art 2D VLA model, and SpatialVLA~\cite{qu2025spatialvla}, a state-of-the-art 3D VLA model. All models use the same number of camera viewpoints, and each task is evaluated with 15 rollouts under consistent test conditions.

\begin{figure}[t]
\includegraphics[width=0.46\textwidth]{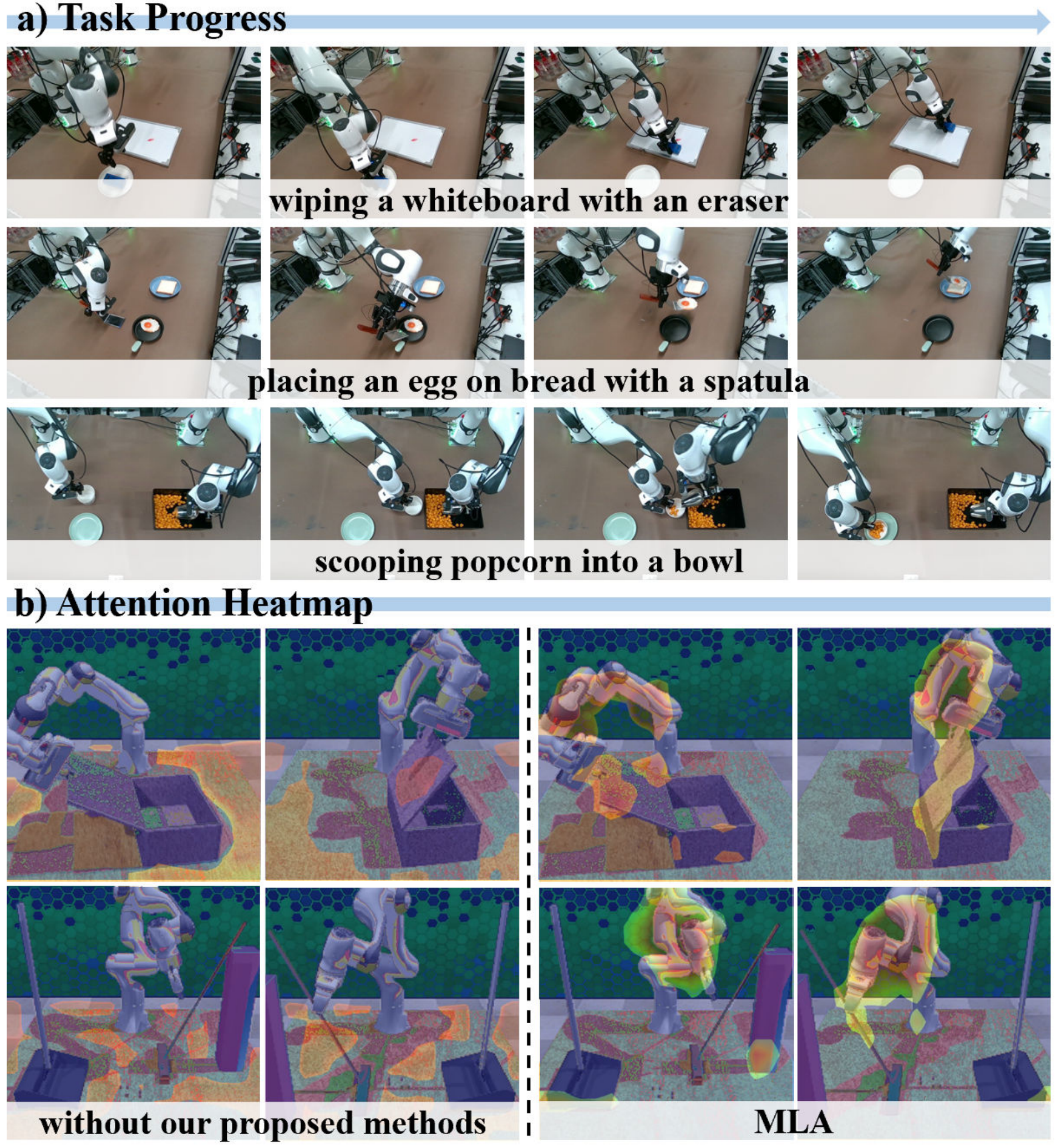}
\centering
\vspace{-0.1cm}
\caption{\textbf{Visualization} of real-world task progress and attention heatmaps from the final-layer output features of MLA.}
\label{fig:progress}
\vspace{-0.45cm}
\end{figure}

\textbf{Results.} As shown in Figure~\ref{fig:realmain}, MLA achieves superior performance across six tasks, outperforming $\pi_0$ and SpatialVLA by an average of 12\% and 24\%, respectively. 
In the Wiping a Whiteboard task, MLA effectively leverages tactile sensing to regulate the downward and lateral movements of the end effector during wiping.
The superior performance is attributed to MLA’s ability to better align with and interpret robotic multisensory inputs, thereby enhancing its perceptual representation of the physical environment compared to the baselines. Furthermore, relative to SpatialVLA, MLA’s advantage arises from its capability to generate future multisensory states, which enables improved modeling of physical dynamics and provides more robust conditions for action generation.
As shown in Figure~\ref{fig:progress}~a), we visualize the robot execution progress for several tasks.

\subsection{Ablation Study}
\label{sec:AS}

To validate each of our proposed contributions, we conducted an ablation study on two real-world tasks, including pressing a stamp onto paper and placing an egg on bread.

\begin{figure}[t]
\includegraphics[width=0.48\textwidth]{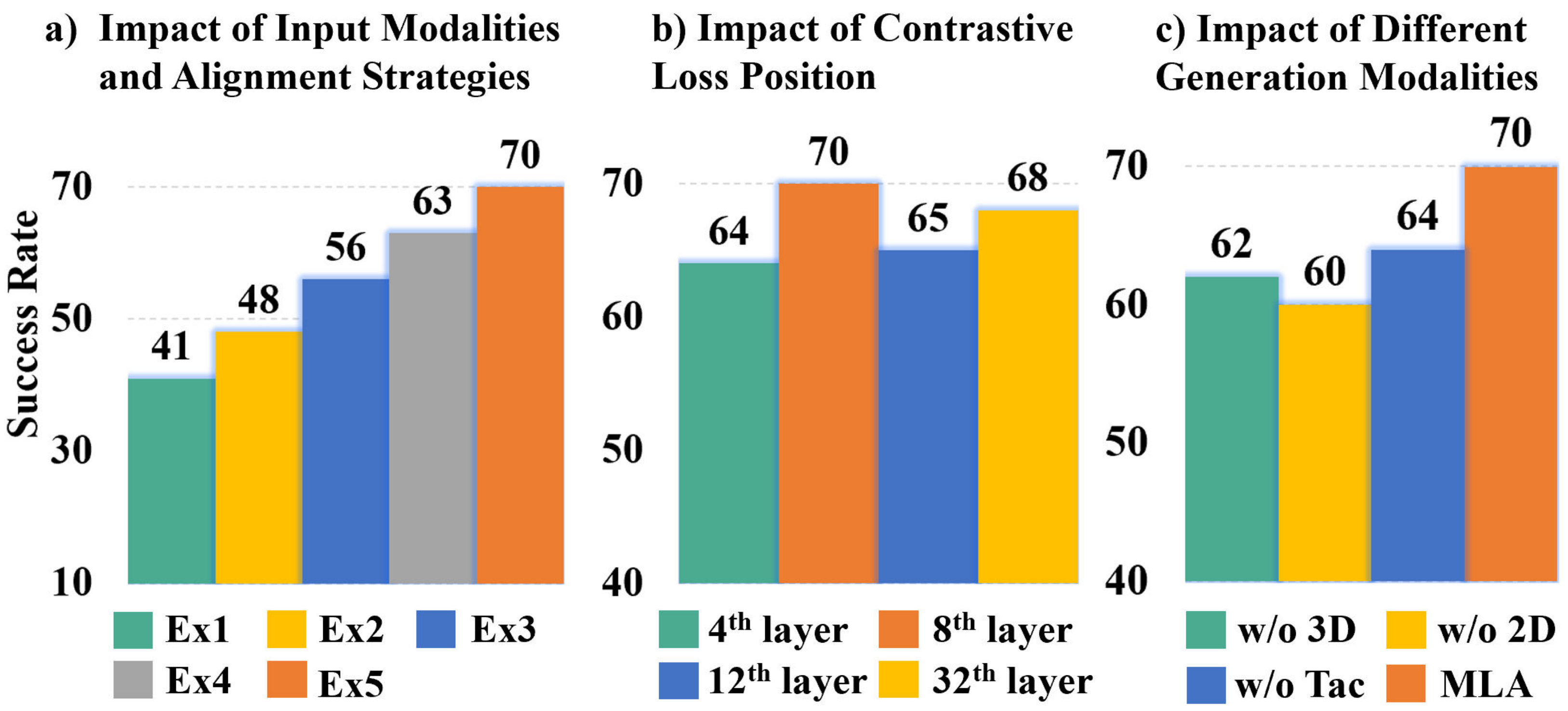}
\centering
\vspace{-0.45cm}
\caption{\textbf{Ablation study.} We systematically analyze the contributions of each component in the MLA model.}
\label{fig:ablation}
\vspace{-0.45cm}
\end{figure}

\textbf{Impact of Input Modalities and Alignment Strategies in the Encoder-Free Multimodal Alignment Scheme.} 
As shown in Figure~\ref{fig:ablation} a), we first examine the role of different input modalities and alignment strategies under the following configurations: (Ex1) 2D image input only, (Ex2) 2D image + 3D point cloud with simple token-level concatenation, (Ex3) 2D image + 3D point cloud + tactile signals with simple token-level concatenation, (Ex4) all modalities with image-level contrastive alignment, and (Ex5) our proposed all modalities with token-level contrastive alignment. Compared with Ex1 and Ex2, Ex3 demonstrates that semantic, spatial, and interactive perception are all critical for contact-rich manipulation. Compared with Ex1–Ex3, Ex5 achieves significant improvements, showing that the proposed position-guided consistency constraint strengthens multimodal representations. Furthermore, compared to Ex4, where multimodal inputs from the same timestep are treated as positives and those from different timesteps as negatives, Ex5 still achieves a 7\% accuracy gain, highlighting the advantage of token-level cross-modal alignment in physical-world perception.

\textbf{Impact of Contrastive Loss Position.} As shown in Figure~\ref{fig:ablation}~b), we investigate the effect of applying contrastive loss at different layers of the LLaMA-2 backbone. Specifically, we select the 4th, 8th, 12th, and 32nd layers for cross-modal alignment during the SFT and post-training stages. The results reveal that applying token-level contrastive learning at the 8th layer yields the best performance, as it aligns features at relatively shallow layers while leaving sufficient subsequent transformer blocks to focus on future state prediction and action generation. Interestingly, applying self-supervision at the 32nd layer yields limited gains, as the final hidden states are already optimized for multiple objectives.

\begin{table*}[t]
    \centering
    \vspace{-0.2cm}
    \caption{\textbf{Results on the RLBench benchmark.} Each model is evaluated over 20 rollouts, with success determined by the built-in RLBench evaluation module. Results report average manipulation success rates (S.R., \%) with variance.}
    \vspace{-0.15cm}
    \resizebox{0.97\textwidth}{!}{ 
    \begin{tabular}{l|cccccccccccc}
        \toprule
        \multirow{2}{*}{Models} & Close & Close & Toilet & Sweep to & Close  & Phone & Take  & Take frame & Place wine & Water  & Mean S.R. \\
                                & box   & laptop lid & seat down & dustpan & fridge & on base & umbrella out & off hanger & at rack & plants & \& Var\\
        \midrule
        OpenVLA~\cite{kim2024openvla}  & 0.60 & 0.35 & 0.75 & 0.55 & 0.85 & 0.20 & 0.30 & 0.15 & 0.20 & 0.05  & 0.40{\color{gray}$\pm$0.02} \\
        $\pi_0$~\cite{black2024pi_0} & 0.85 & \textbf{0.95} & 0.90 & 0.85 & \textbf{1.00} & 0.05 & 0.10 & \textbf{0.90} & 0.65 & 0.25 & 0.65{\color{gray}$\pm$0.04}\\
        HybridVLA~\cite{liu2025hybridvla}  & 0.85 & 0.75 & \textbf{1.00} & 0.80 & 0.95 & 0.50 & 0.50 & 0.30 & 0.70 & 0.25 & 0.66{\color{gray}$\pm$0.05}\\
        SpatialVLA~\cite{qu2025spatialvla} & 0.80 & 0.70 & 0.85 & 0.20 & 0.80 & 0.15 & 0.25 & 0.40 & 0.15 & 0.30 & 0.46{\color{gray}$\pm$0.03} \\
        UP-VLA~\cite{zhang2025up} & 0.80 & 0.40 & 0.65 & 0.10 & 0.80 & 0.15 & 0.35 & 0.55 & 0.20 & 0.20 & 0.42{\color{gray}$\pm$0.04} \\
        \textit{DreamVLA*}~\cite{zhang2025dreamvla} & \textbf{0.95} & 0.75 & 0.95 & 0.25 & \textbf{1.00} & 0.35 & \textbf{0.55} & 0.50 & \textbf{0.85} & 0.35 & 0.65{\color{gray}$\pm$0.05} \\
        \textbf{MLA} & \textbf{0.95} & 0.90 & \textbf{1.00} & \textbf{1.00} & 0.95 & \textbf{0.60} & 0.50 & \textbf{0.90} & 0.75 & \textbf{0.55} & \textbf{0.81{\color{gray}$\pm$0.03}} \\
        \bottomrule
    \end{tabular}}
    \label{tab:sim}
    \vspace{-0.4cm}
\end{table*}

\textbf{Impact of Multimodal Data Encoding Methods.} We also compare injecting additional 2D~\cite{zhai2023siglip} and 3D~\cite{jia2025lift3d} with our proposed approach that repurposes the LLM itself as a unified perception module. The results show that introducing extra encoders not only decreases performance (–7\%) but also reduces inference efficiency.

\textbf{Impact of Different Generation Modalities in Future State Generation.} As shown in Figure~\ref{fig:ablation} c), building upon the MLA model following SFT, we further evaluated three ablation variants during post-training: (1) without image generation, (2) without point cloud generation, and (3) without tactile signal generation. The results indicate that removing future state generation from any modality leads to a drop in accuracy, reaffirming that generating comprehensive semantic, spatial, and interactive information provide more robust feature conditions for action generation.
Finally, we investigate the impact of predicting \textbf{future adjacent frames (64\%)} versus \textbf{future keyframes (70\%)} on manipulation performance. The results show that predicting adjacent frames introduces high redundancy, leading to limited improvements in motion planning and dynamic representation of MLA.
\vspace{-0.1cm}
\begin{table}[h]
\centering
\caption{\textbf{Generalization experiments.} 
Visualization of the two generalization scenarios along with the corresponding quantitative results.
The red boxes highlight the differences from the training setup.}
\begin{minipage}{0.48\textwidth}
    \includegraphics[width=\linewidth]{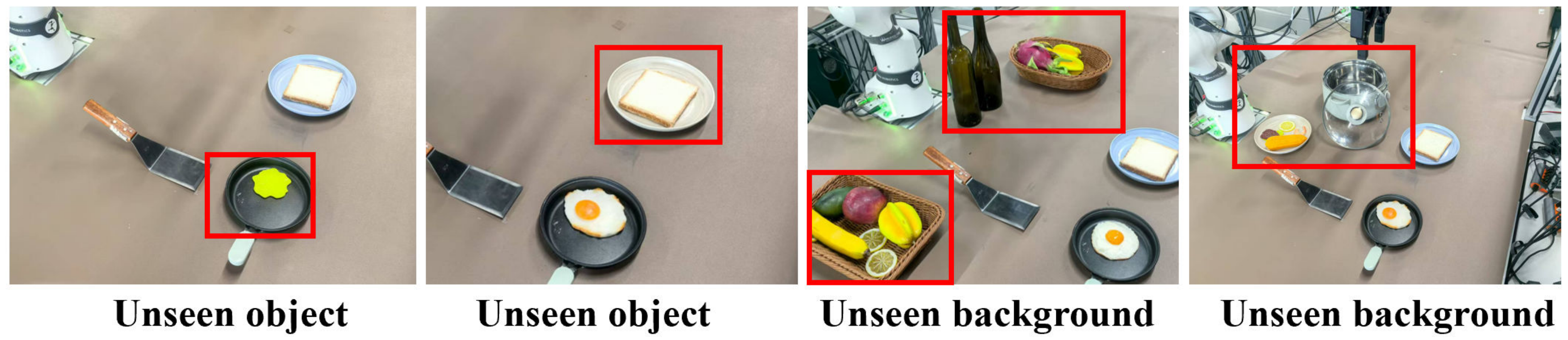}
\end{minipage}\hfill
\vspace{0.1cm}
\begin{minipage}{0.48\textwidth}
    \centering
    \begin{tabular}{l|ccc}
        \toprule
        Model   & Original & Unseen Object & Unseen Background \\
        \midrule
         $\pi_0$  & 47 & 35 (-26\%) & 25 (-47\%)\\
        MLA  & 53 & 45 (-15\%) & 40 (-25\%) \\
        \bottomrule
    \end{tabular}
\end{minipage}
\vspace{-0.35cm}
\label{tab:gene}
\end{table}

\subsection{Generalization Experiment}
\label{sec:GE}

As shown in Table~\ref{tab:gene}, we designed two common generalization scenarios to compare our MLA with $\pi_0$, including unseen manipulated objects and unseen complex backgrounds. The most challenging task, placing an egg on bread with a spatula, is selected as the evaluation task. For unseen objects, we replace the egg with lettuce and change the color of the target plate. Across these foreground modifications, MLA shows almost no decrease in success rate for the initial subtask. For unseen backgrounds, cluttered scenes are introduced during testing by adding unseen objects around the manipulated object. Even under such challenging background conditions, MLA maintains a 40\% success rate in completing the entire task. These results demonstrate that MLA can better perceive and reason about robotic manipulation scenes, whether facing semantic variations in manipulated objects or background interference. This robustness is attributed to its strong multimodal perception capability and its ability to anticipate the future states of manipulated objects.

\subsection{Simulation Experiment}
\label{sec:SE}

\textbf{Simulation Benchmark.} To systematically evaluate the performance of MLA, we conducted experiments on 10 tasks in the RLBench~\cite{james2020rlbench} benchmark, which is based on the CoppeliaSim simulator. For each task, 100 demonstration trajectories were collected using the official Motion Planning Library~\cite{sucan2012the-open-motion-planning-library}. Observations consist of a front-view camera image and the corresponding point cloud data. We extract the keyframes following the approach in~\cite{jia2025lift3d}.

\textbf{Training and Evaluation Details.} 
As tactile sensing in simulation is not realistic, we provide only image and point cloud modalities to MLA and all baseline methods. We selected several state-of-the-art baselines from relevant domains, including OpenVLA~\cite{kim2024openvla}, $\pi_0$~\cite{black2024pi_0}, HybridVLA~\cite{liu2025hybridvla}, SpatialVLA~\cite{qu2025spatialvla}, UP-VLA~\cite{zhang2025up}, and $DreamVLA^{*}$~\cite{zhang2025dreamvla}. For each baseline, we loaded the officially released pretrained checkpoints. Since $DreamVLA^{*}$ does not provide a general large-scale pretrained checkpoint, we re-implemented its input and generation strategy on our backbone for a fair comparison. All tasks were trained jointly, and evaluation was performed using 20 rollouts per task.

\textbf{Results.} As shown in Table~\ref{tab:sim}, MLA achieves an average score of 81\% across 10 tasks, significantly outperforming $\pi_0$ (65\%), SpatialVLA (46\%), and other baselines. The improvements are particularly notable on more challenging tasks, such as \textit{Place Wine at Rack Location} and \textit{Water plants}. These results validate the effectiveness of our proposed multimodal alignment and future multisensory generation post-training, which enable MLA to progressively enhance its representations and achieve higher action accuracy. They also demonstrate that our paradigm remains effective even without access to expensive sensors such as tactile devices.
Furthermore, as shown in Figure~\ref{fig:progress}~b), we visualize the attention heatmaps from the output features of MLA and a variant without our proposed approach. The results clearly show that MLA learns better feature representations and focuses more effectively on both the robot and the manipulated objects.

\section{CONCLUSIONS}

In this work, we introduced MLA, a multisensory language–action model that integrates 2D visual, 3D geometric, and tactile cues through encoder-free multimodal alignment and enhances physical-world understanding via future multisensory generation. We progressively equip a LLM with the ability to integrate perception, understanding, and action generation from multisensory inputs in the real world through large-scale pretraining, supervised fine-tuning, and post-training. MLA not only achieves state-of-the-art performance and demonstrates strong generalization across both real-world and simulation tasks, but also provides a new multimodal foundation model paradigm for the community.

\section{Acknowledgement}
This work was supported by the National Natural Science Foundation of China (625B2007).
This work was also supported by the National Natural Science Foundation of China (62476011).
This work was also supported by Beijing Innovation Center of Humanoid Robotics.



{
\bibliographystyle{IEEEtran}
\bibliography{IEEEabrv,reference}
}

\newpage

\appendix

\subsection{Real-world Robot Set-up}
\label{sec:RRS}

\begin{figure}[h]
\vspace{-0.2cm}
\includegraphics[width=0.36\textwidth]{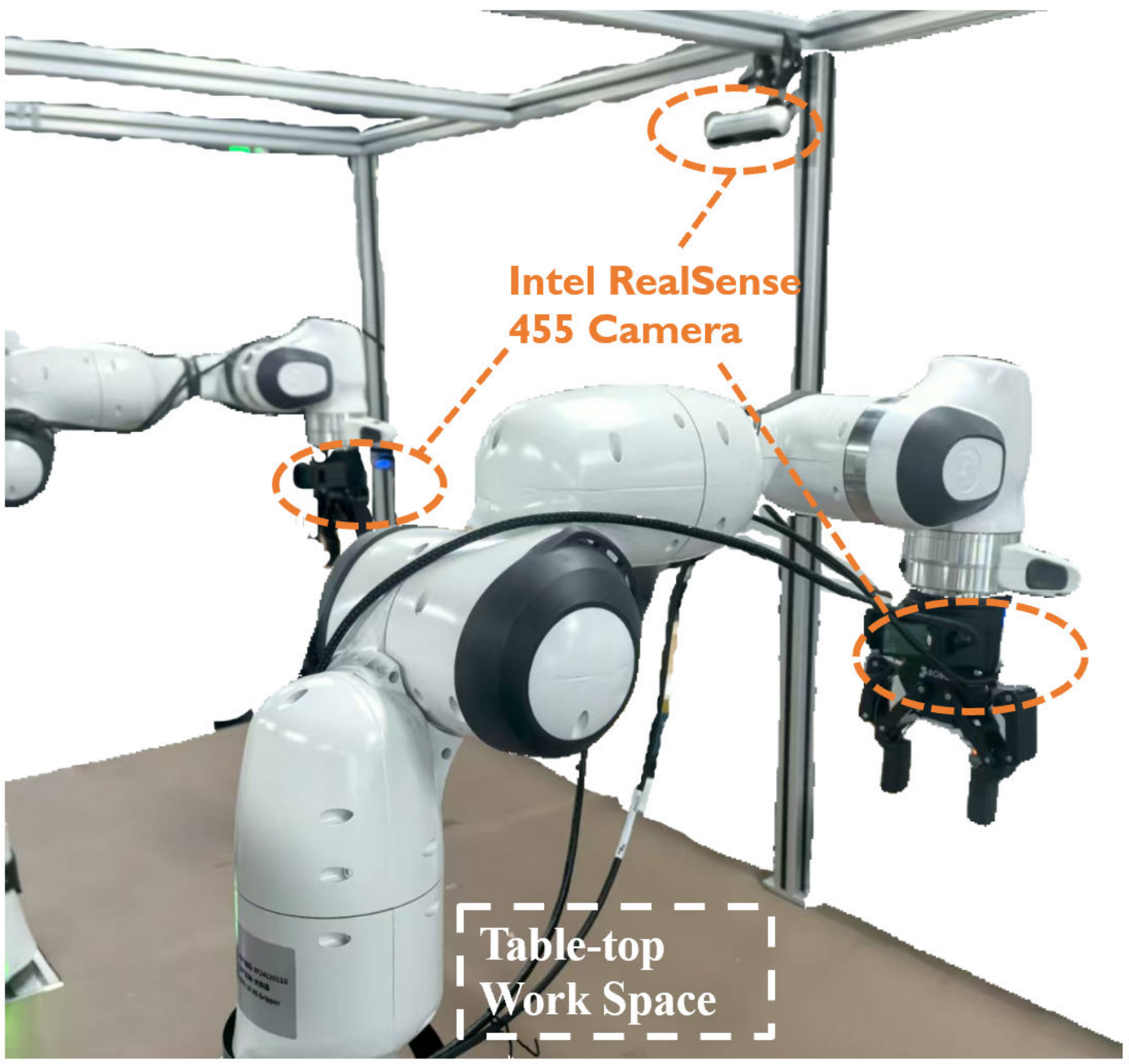}
\centering
\vspace{-0.05cm}
\caption{\textbf{Dual-arm Experiment Setup.} For dual-arm experiments, we use two parallel FR3 arms to achieve collaborative manipulation.}
\label{fig:dualset}
\vspace{-0.35cm}
\end{figure}

The experimental setup and environments for the single-arm and dual-arm setups are shown in Figure~\ref{fig:singleset} and Figure~\ref{fig:dualset}, respectively.

For single-arm tasks, we employ a Franka Research 3 robotic arm equipped with a ROBOTIQ adaptive gripper as the end-effector. Visual observations are provided by two Intel RealSense D455 cameras, one positioned at a right-front third-person viewpoint and the other mounted on the wrist. In addition, two Tashan TS-E-A tactile sensors are attached to the fingertips of the gripper to capture tactile feedback.

For dual-arm tasks, we utilize two parallel Franka Research 3 arms with the same end-effector configuration as the single-arm asset. The observation setup includes an additional front-facing RealSense D455 camera along with two wrist-mounted cameras, ensuring comprehensive multi-view perception.

\subsection{Additional Dataset Details}
\label{sec:ADD}

\begin{table}[h]
\centering
\caption{\textbf{Datasets used for pre-training.} 
The names of selected 28 datasets for large-scale pretraining and their sampling ratios (\%).}
\begin{tabular}{l|r}
    \toprule
    Dataset & Ratio (\%) \\
    \midrule
    Austin Buds~\cite{zhu2022bottom} & 0.01 \\
    Austin Sailor~\cite{nasiriany2022sailor} & 0.04 \\
    Austin Sirius~\cite{liu2022robot} & 0.10 \\
    BC-Z~\cite{jang2022bc} & 7.54 \\
    Berkeley Autolab Ur5~\cite{BerkeleyUR5Website} & 0.35 \\
    BridgeV2~\cite{ebert2022bridge,walke2023bridgedata} & 20.93 \\
    CMU Stretch~\cite{mendonca2023structured} & 0.02 \\
    DLR Sara Grid Clamp~\cite{padalkar2023guided} & 0.02 \\
    DROID~\cite{khazatsky2024droid} & 4.82 \\
    Dobb-E~\cite{shafiullah2023dobbe} & 0.18 \\
    FMB Dataset~\cite{luo2023fmb}& 1.50 \\
    Fractal~\cite{rt12022arxiv} & 13.67 \\
    Furniture Bench~\cite{heo2023furniturebench} & 0.09 \\
    Jaco Play~\cite{dass2023jacoplay} & 0.19 \\
    Kuka~\cite{kalashnikov2018qt} & 20.22 \\
    Language Table~\cite{lynch2023interactive} & 7.70 \\
    Maniskill~\cite{gu2023maniskill2unifiedbenchmarkgeneralizable} & 5.26 \\
    Nyu Franka Play~\cite{cui2022play} & 0.24 \\
    Robo-Net~\cite{dasari2020robonet} & 11.53 \\
    Roboset~\cite{kumar2023robohive} & 3.18 \\
    RoboTurk~\cite{DBLP:journals/corr/abs-1811-02790} & 0.70 \\
    Stanford Hydra~\cite{belkhale2023hydra} & 0.20 \\
    Taco Play~\cite{rosetebeas2022latent,mees2023grounding} & 1.26 \\
    Toto~\cite{zhou2023train} & 0.17 \\
    Utokyo Pr2 Fridge~\cite{oh2023pr2utokyodatasets} & 0.01 \\
    Utokyo Pr2 Tabletop~\cite{oh2023pr2utokyodatasets} & 0.04 \\
    Utokyo Xarm Pap~\cite{matsushima2023weblab} & 0.04 \\
    RoboMIND~\cite{wu2024robomind} & 0.10 \\
    \bottomrule
\end{tabular}
\label{tab:datasets}
\end{table}

\textbf{Large-scale Pretraining Dataset.} To ensure the quality and consistency of training data, we curated 28 high-quality datasets from Open-X-Embodiment~\cite{open_x_embodiment_rt_x_2023}, DROID~\cite{khazatsky2024droid} and RoboMIND~\cite{wu2024robomind} datasets and applied customized sampling ratios, resulting in a total of 570K trajectories and 36M frames, as shown in Table~\ref{tab:datasets}. The action representations across datasets were unified to align with those used in the fine-tuning stage, thereby maximizing the utility of pretraining. During pretraining, since these datasets only provide 2D image observations, we restricted the input modalities to 2D RGB images, language instructions, and robot states. Meanwhile, the token sequences corresponding to 3D point clouds and tactile signals were reserved as empty tokens, ensuring consistency of input sequences between pretraining and fine-tuning. In the fine-tuning stage, we further incorporated multi-view images, which were encoded through a shared tokenizer and concatenated sequentially after the single-view image tokens if needed.

\textbf{Self-collected and Real-world Dataset.} Building upon our real-world experimental setup, we present the detailed procedures of six robotic tasks, while the four tasks:

\textit{1. Pressing a stamp onto paper.}  
This task requires the robot to first pick up a stamp, and, once a stable grasp is established, press it downward to imprint on a sheet of paper. The stamping process involves significant tangential force feedback, and the completion of stamping can be detected by the abrupt change in tangential force, while visual feedback is not involved.

\textit{2. Wiping a whiteboard with an eraser.}
This task requires the robot to first pick up a whiteboard eraser and then erase the colored blocks on the board by recognizing them through vision. In addition to the visual feedback used for judging the erasure, tactile signals during lateral movements also provide cues about the distance traveled and whether contact with the board has been established.

\textit{3. Placing a dish on a rack.}
This task requires the robot to first grasp a plate from its edge and then perform a large rotational motion to accurately place it into a dish rack. Since the plate must be inserted in an upright orientation, the model is required to form an accurate spatial perception and precisely predict the necessary rotation.

\textit{4. Placing an egg on bread with a spatula.}
This task can be divided into three subtasks: picking up the spatula, lifting the egg, and placing the egg onto the bread. The subtask of lifting the egg requires the model to perceive the contact between the spatula tip and the pan, and to slide along the bottom of the egg to lift it. Moreover, this process demands precise position prediction. Overall, the task heavily relies on spatial perception and tactile feedback.

\textit{5. Scooping popcorn into a bowl.} 
In this task, the robot’s right arm uses a spatula to scoop popcorn from a box and pour it into a bowl held by the left arm. The scooping process relies on tactile feedback, while the pouring step requires spatial coordination between the two arms.

\begin{figure}[!t]
\vspace{-0.2cm}
\includegraphics[width=0.48\textwidth]{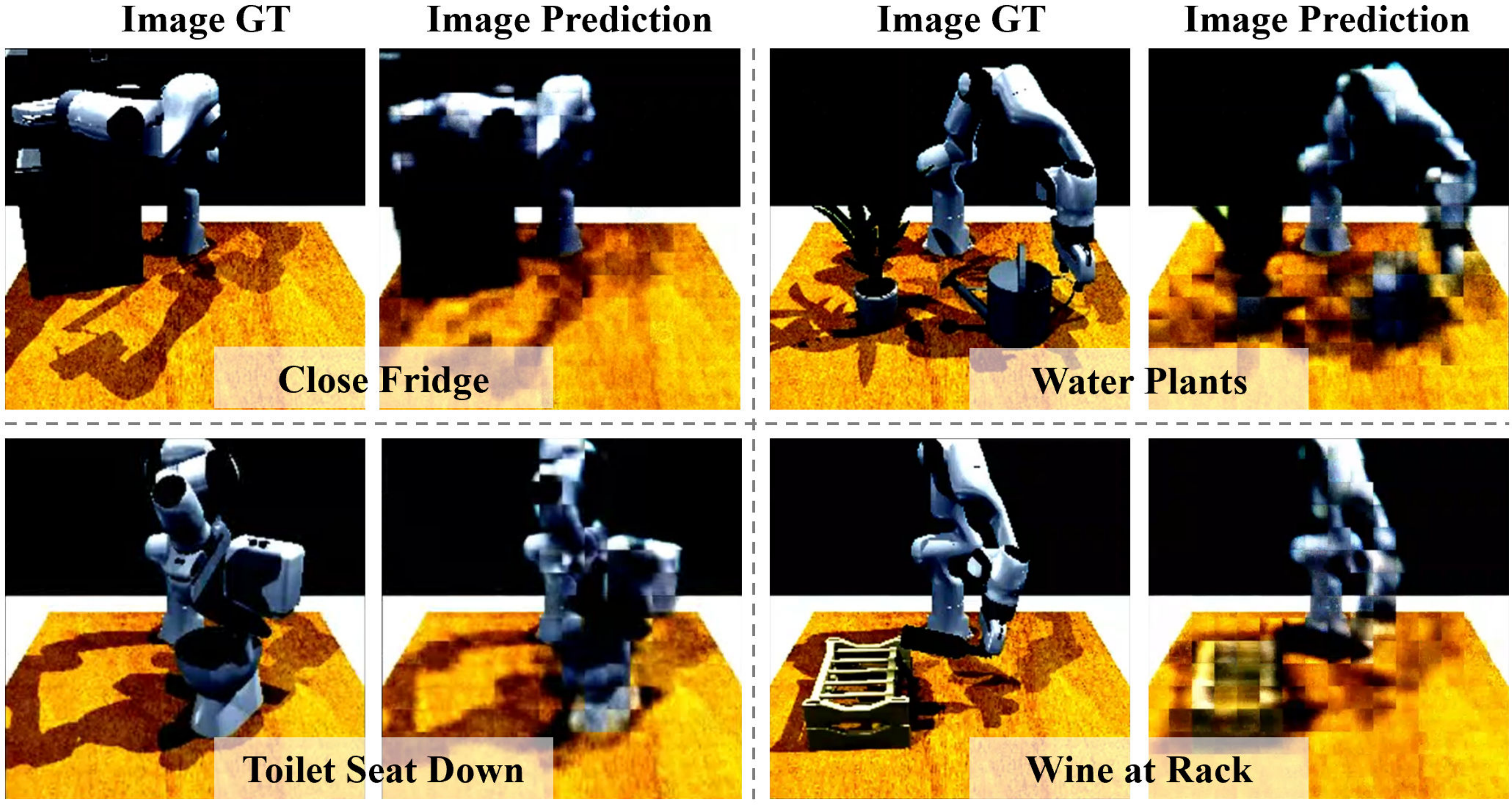}
\centering
\vspace{-0.05cm}
\caption{\textbf{Visualization of Image Prediction.} We selected four image prediction results during training. For each sample, the image ground truth is shown on the left and the prediction result is shown on the right.}
\label{fig:imgpred}
\vspace{-0.35cm}
\end{figure}

\textit{6. Opening a pot lid and picking corn from the pot.}
In this task, the left arm first opens the pot lid, after which the right arm retrieves corn from the pot and places it onto a plate, followed by the left arm closing the lid. Each step requires precise position prediction, and the two arms must collaborate to avoid collisions and other undesired interactions.

\subsection{Additional Visualizations}
\label{sec:ADV}

In this section, we provide additional visualizations of the multimodel future generation results during training on RLBench data.

\begin{figure}[h]
\vspace{-0.2cm}
\includegraphics[width=0.48\textwidth]{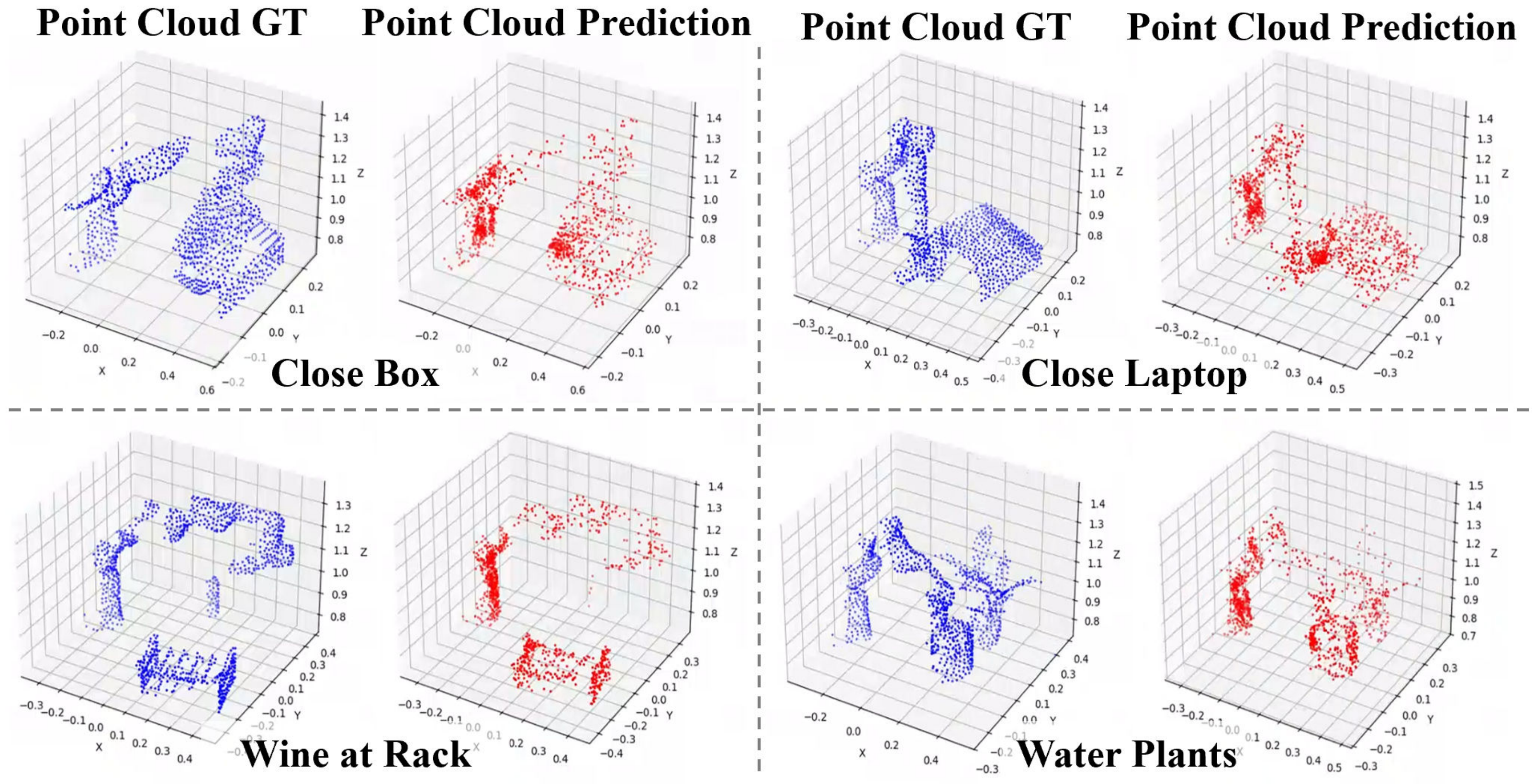}
\centering
\vspace{-0.05cm}
\caption{\textbf{Visualization of Point Cloud Prediction.} We selected four point cloud prediction results during training. For each sample, the point cloud ground truth is shown on the left and the prediction result is shown on the right.}
\label{fig:pcpred}
\vspace{-0.35cm}
\end{figure}

\textbf{Image Prediction.} As illustrated in Figure \ref{fig:imgpred}, MLA is supervised during training by performing image-to-image prediction with the next key frame, thereby forecasting the future image observation. In the context of multi-task joint training, MLA is able to predict most of the critical scene information, including the robot arm’s pose, the state of the manipulated object, and the lighting conditions simulating real-world environments. This enhances the model’s ability to perceive the overall scene dynamics.

\textbf{Point Cloud Prediction.} As shown in Figure \ref{fig:pcpred}, during the multimodal future prediction process, MLA simultaneously receives supervision from the next key frame’s point cloud. By employing the supervision method described in Section \ref{sec:FMG}, MLA efficiently predicts point clouds with accurate geometric features and rich local information. The concurrent prediction of both image and point cloud modalities fully leverages the unified representation within the MLA model. Furthermore, through predicting the point cloud modality, the model’s understanding of object spatial positions is significantly enhanced.

\end{document}